\documentclass[10pt,twocolumn,letterpaper]{article}

\usepackage{cvpr}
\usepackage{times}
\usepackage{epsfig}
\usepackage{graphicx}
\usepackage{amsmath}
\usepackage{amssymb}
\usepackage{verbatim}

\usepackage{tabularx}
\usepackage{makecell}
\usepackage{authblk}

\usepackage[pagebackref=true,breaklinks=true,letterpaper=true,colorlinks,bookmarks=false]{hyperref}

\cvprfinalcopy % *** Uncomment this line for the final submission
 
\def\cvprPaperID{****} % *** Enter the CVPR Paper ID here

\ifcvprfinal\fi

\begin{document}

%%%%%%%%% TITLE
\title{Deep Alignment Network: A convolutional neural network for robust face alignment}

\author{Marek Kowalski}
\author{Jacek Naruniec}
\author{Tomasz Trzcinski}
\affil {Warsaw University of Technology}
\affil {\tt\small m.kowalski@ire.pw.edu.pl, j.naruniec@ire.pw.edu.pl, t.trzcinski@ii.pw.edu.pl}

\maketitle
\thispagestyle{empty}

%%%%%%%%% ABSTRACT
\begin{abstract}
In this paper, we propose Deep Alignment Network (DAN), a robust face alignment method based on a deep neural network architecture. DAN consists of multiple stages, where each stage improves the locations of the facial landmarks estimated by the previous stage. Our method uses entire face images at all stages, contrary to the recently proposed face alignment methods that rely on local patches. This is possible thanks to the use of {\it landmark heatmaps} which provide visual information about landmark locations estimated at the previous stages of the algorithm. The use of entire face images rather than patches allows DAN to handle face images with large variation in head pose and difficult initializations. An extensive evaluation on two publicly available datasets shows that DAN reduces the state-of-the-art failure rate by up to 70\%. Our method has also been submitted for evaluation as part of the Menpo challenge.
\end{abstract} 

\section{Introduction}
The goal of face alignment is to localize a set of predefined facial landmarks (eye corners, mouth corners etc.) in an image of a face. Face alignment is an important component of many computer vision applications, such as face verification \cite{DeepFace}, facial emotion recognition \cite{FacialAffect}, human-computer interaction \cite{HCI} and facial motion capture \cite{Animation}.

Most of the face alignment methods introduced in the recent years are based on shape indexed features \cite{SDM, LBF, ESR, CFSS, MDM}. In these approaches image features, such as SIFT \cite{SDM, CFSS} or learned features \cite{LBF, MDM}, are extracted from image patches extracted around each of the landmarks. The features are then used to iteratively refine the estimates of landmark locations. While those approaches can be successfully applied to face alignment in many photos, their performance on the most challenging datasets~\cite{300-W} leaves room for improvement~\cite{ESR, SDM, CFSS, MDM}. We believe that this is due to the fact that for the most difficult images the features extracted at disjoint patches do not provide enough information and can lead the method into a local minimum.

In this work, we address the above shortcoming by proposing a novel face alignment method which we dub Deep Alignment Network (DAN). It is based on a multi-stage neural network where each stage refines the landmark positions estimated at the previous stage, iteratively improving the landmark locations. The input to each stage of our algorithm (except the first stage) are a face image normalized to a canonical pose and an image learned from the dense layer of the previous stage. To make use of the entire face image during the process of face alignment, we additionally input at each stage a {\it landmark heatmap}, which is a key element of our system.

\begin{figure*}[tb]
\centering
\includegraphics[width=0.9\textwidth]{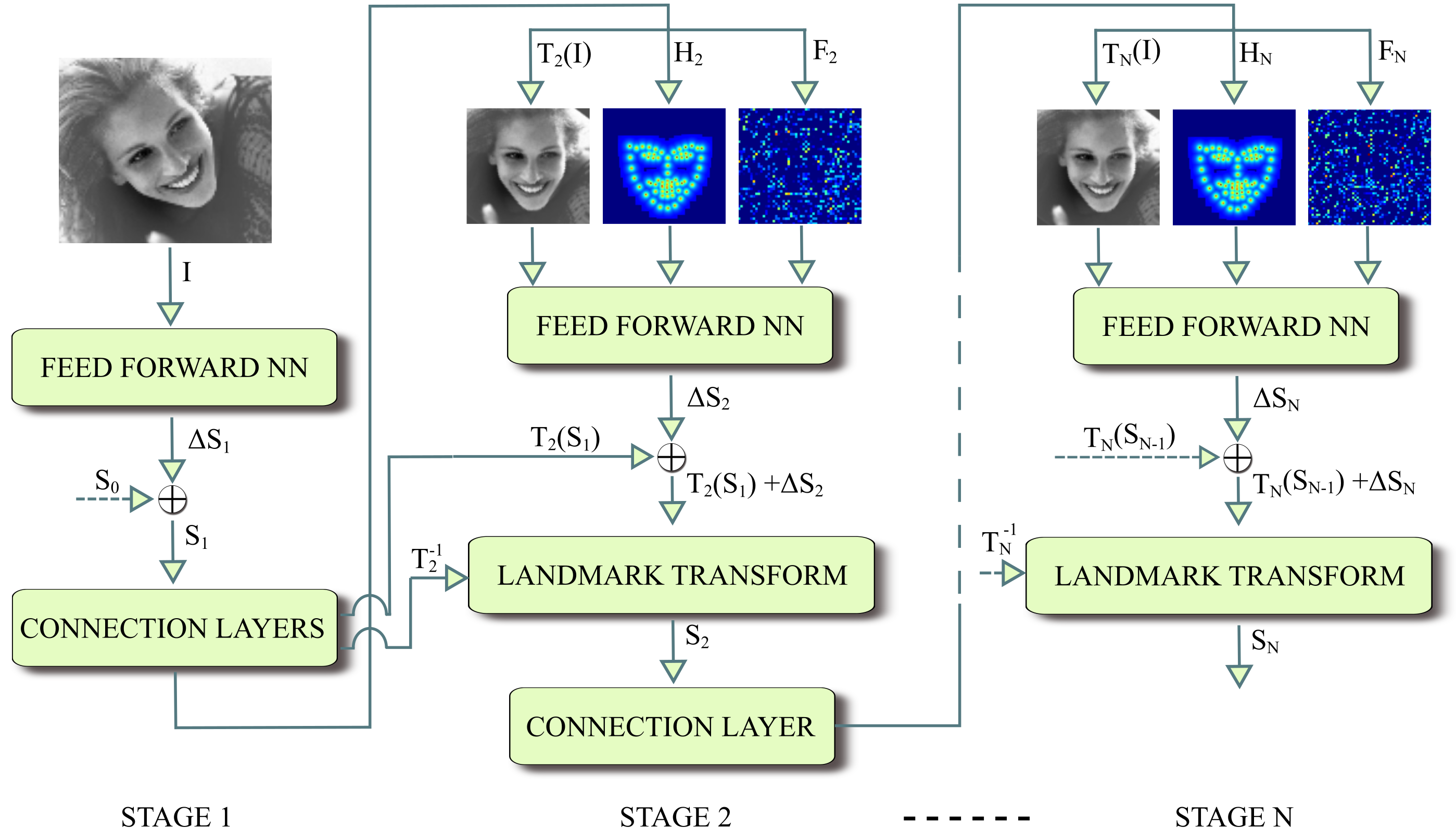}
\caption{A diagram showing an outline of the proposed method. Each stage of the neural network refines the landmark location estimates produced by the previous stage, starting with an initial estimate $S_0$. The connection layers form a link between the consecutive stages of the network by producing the landmark heatmaps $H_t$, feature images $F_t$ and a transform $T_t$ which is used to warp the input image to a canonical pose. By introducing landmark heatmaps and feature images we can transmit crucial information, including the landmark location estimates, between the stages of our method. }
 \label{fig:diagram1}
\end{figure*}

A landmark heatmap is an image with high intensity values around landmark locations where intensity decreases with the distance from the nearest landmark. The convolutional neural network can use the heatmaps to infer the current estimates of landmark locations in the image and thus refine them. An example of a landmark heatmap can be seen in Figure~\ref{fig:diagram1} which shows an outline of our method. By using landmark heatmaps, our DAN algorithm is able to reduce the failure rate on the 300W public test set by a large margin of 72\% with respect to the state of the art.

To summarize, the three main contributions of this work are the following:
\begin{enumerate}
  \item We introduce landmark heatmaps which transfer the information about current landmark location estimates between the stages of our method. This improvement allows our method to make use of the entire image of a face, instead of local patches, and avoid falling into local minima.
  \item The resulting robust face alignment method we propose in this paper reduces the failure rate by 60\% on the 300W private test set \cite{300-W} and 72\% on the 300-W public test set \cite{300-W} compared to the state of the art.
  \item Finally, we publish both the source code of our implementation of the proposed method and the models used in the experiments. 
\end{enumerate}

The remainder of the paper is organized in the following manner. In section \ref{related} we give an overview of the related work. In section \ref{DAN} we provide a detailed description of the proposed method. Finally, in section \ref{experiments} we perform an evaluation of DAN and compare it to the state of the art. 

\section{Related work} \label{related}
Face alignment has a long history, starting with the early Active Appearance Models \cite{AAM, AAM-revisited}, moving to Constrained Local Models \cite{CLM1, CLM2} and recently shifting to methods based on Cascaded Shape Regression (CSR) \cite{SDM, ESR, LBF, cGPRT, KRFWS, MIX, Trees} and deep learning \cite{Zhou2013, Fan2016, MDM, RAR, Bulat16}.

In CSR based methods, the face alignment begins with an initial estimate of the landmark locations which is then refined in an iterative manner. The initial shape $S_0$ is typically an average face shape placed in the bounding box returned by the face detector\cite{ESR, SDM, LBF, MDM}. Each CSR iteration is characterized by the following equation:

\begin{equation}
S_{t+1}=S_t + r_t(\phi(I, S_{t})),
\end{equation}
where $S_t$ is the estimate of landmark locations at iteration $t$, $r_t$ is a regression function which returns the update to $S_t$ given a feature $\phi$ extracted from image $I$ at the landmark locations. 

The main differences between the variety of CSR based methods introduced in the literature lie in the choice of the feature extraction method $\phi$ and the regression method $r_t$. For instance, Supervised Descent Method (SDM) \cite{SDM} uses SIFT \cite{SIFT} features and a simple linear regressor. LBF \cite{LBF} takes advantage of sparse features generated from binary trees and intensity differences of individual pixels. LBF uses Support Vector Regression \cite{SVR} for regression which, combined with the sparse features, leads to a very efficient method running at up to 3000 fps. 

Coarse to Fine Shape Searching (CFSS) \cite{CFSS}, similarly to SDM, uses SIFT features extracted at landmark locations. However the regression step of CSR is replaced with a search over the space of possible face shapes which goes from coarse to fine over several iterations. This reduces the probability of falling into a local minimum and thus improves convergence.

MIX \cite{MIX} also uses SIFT for feature extraction, while regression is performed using a mixture of experts, where each expert is specialized in a certain part of the space of face shapes. Moreover MIX, warps the input image before each iteration so that the current estimate of the face shape matches a predefined canonical face shape. 

Mnemonic Descent Method (MDM) \cite{MDM} fuses the feature extraction and regression steps of CSR into a single Recurrent Neural Network that is trained end-to-end. MDM also introduces memory into the process which allows information to be passed between CSR iterations.

While all of the above mentioned methods perform face alignment based only on local patches, there are some methods \cite{Zhou2013,Fan2016} that estimate initial landmark positions using the entire face image and use local patches for refinement. In contrast, DAN localizes the landmarks based on the entire face image at all of its stages.

The use of heatmaps for face alignment related tasks precedes the proposed method. One method that uses heatmaps is \cite{Bulat16}, where a neural network outputs predictions in the form of a heatmap. In contrast, the proposed method uses heatmaps solely as a means for transferring information between stages.

The development of novel methods contributes greatly in advancing face alignment. However it cannot be overlooked that the publication of several large scale datasets of annotated face images \cite{300-W, Reannotation} also had a crucial role in both improving the state of the art and the comparability of face alignment methods. 

\section{Deep Alignment Network}  \label{DAN}
In this section, we describe our method, which we call the Deep Alignment Network (DAN). DAN is inspired by the Cascade Shape Regression (CSR) framework, just like CSR our method starts with an initial estimate of the face shape $S_0$ which is refined over several iterations. However, in DAN we substitute each CSR iteration with a single stage of a deep neural network which performs both feature extraction and regression. The major difference between DAN and approaches based on CSR is that DAN extracts features from the entire face image rather than the patches around landmark locations. This is achieved by introducing additional input to each stage, namely a landmark heatmap which indicates the current estimates of the landmark positions within the global face image and transmits this information between the stages of our algorithm. An outline of the proposed method is shown in Figure \ref{fig:diagram1}.

Therefore, each stage of DAN takes three inputs: the input image $I$ which has been warped so that the current landmark estimates are aligned with the canonical shape $S_0$, a landmark heatmap $H_t$ and a feature image $F_t$ which is generated from a dense layer connected to the penultimate layer of the previous stage $t-1$. The first stage only takes the input image as the initial landmarks are always assumed to be the average face shape $S_0$ located in the middle of the image.

\begin{figure}[tb]
\centering
\includegraphics[width=0.9\linewidth]{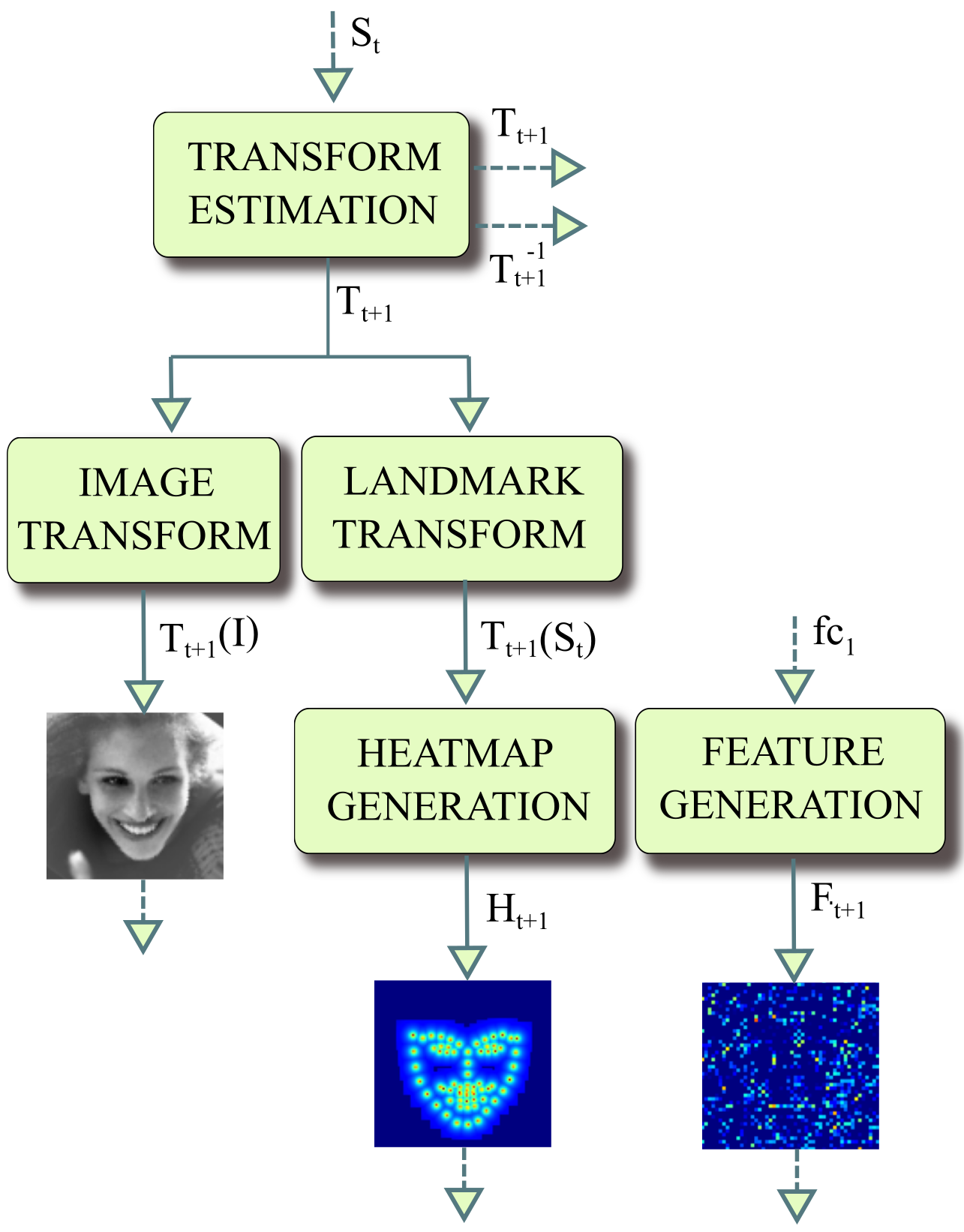}
\caption{A diagram showing an outline of the connection layers. The landmark locations estimated by the current stage $S_{t}$ are first used to estimate the normalizing transform $T_{t+1}$ and its inverse $T^{-1}_{t+1}$. $T_{t+1}$ is subsequently used to transform the input image $I$ and $S_{t}$. The transformed shape $T_{t+1}(S_t)$ is then used to generate the landmark heatmap $H_{t+1}$. The feature image $F_{t+1}$ is generated using the fc1 dense layer of the current stage $t$.}
 \label{fig:diagramconnection}
\end{figure}

A single stage of DAN consists of a feed-forward neural network which performs landmark location estimation and \textit{connection layers} that generate the input for the next stage. The details of the feed-forward network are described in subsection \ref{sec:feed-forward}. The connection layers consist of the Transform Estimation layer, the Image Transform layer, Landmark Transform layer, Heatmap Generation layer and Feature Generation layer. The structure of the connection layers is shown in Figure \ref{fig:diagramconnection}. 
 
The transform estimation layer generates the transform $T_{t+1}$, where $t$ is the number of the stage. The transformation is used to warp the input image $I$ and the current landmark estimates $S_{t}$ so that $S_{t}$ is close to the canonical shape $S_0$. The transformed landmarks $T_{t+1}(S_{t})$ are passed to the heatmap generation layer. The inverse transform $T^{-1}_{t+1}$ is used to map the output landmarks of the consecutive stage back into the original coordinate system. 

The details of the Transform Estimation, Image Transform and Landmark Transforms layer are described in subsection \ref{sec:normalization}. The Heatmap Generation and Feature Image layers are described in sections \ref{sec:heatmap}, \ref{sec:feature}. Section \ref{sec:training} details the training procedure.

\begin{table}
\centering
\caption{ Structure of the feed-forward part of a Deep Alignment Network stage. The kernels are described as height $\times$ width $\times$ depth, stride.} \label{tab:feed-forward}
\begin{tabular}{| c | c | c | c |}
\hline
Name & Shape-in & Shape-out & Kernel \\
\hline
conv1a & 112$\times$112$\times$1 & 112$\times$112$\times$64 & 3$\times$3$\times$1,1\\
\hline
conv1b & 112$\times$112$\times$64 & 112$\times$112$\times$64 & 3$\times$3$\times$64,1\\
\hline
pool1 & 112$\times$112$\times$64 & 56$\times$56$\times$64 & 2$\times$2$\times$1,2\\
\hline
conv2a & 56$\times$56$\times$64 & 56$\times$56$\times$128 & 3$\times$3$\times$64,1\\
\hline
conv2b & 56$\times$56$\times$128 & 56$\times$56$\times$128 & 3$\times$3$\times$128,1\\
\hline
pool2 & 56$\times$56$\times$128 & 28$\times$28$\times$128 & 2$\times$2$\times$1,2\\
\hline
conv3a & 28$\times$28$\times$128 & 28$\times$28$\times$256 & 3$\times$3$\times$128,1\\
\hline
conv3b & 28$\times$28$\times$256 & 28$\times$28$\times$256 & 3$\times$3$\times$256,1\\
\hline
pool3 & 28$\times$28$\times$256 & 14$\times$14$\times$256 & 2$\times$2$\times$1,2\\
\hline
conv4a & 14$\times$14$\times$256 & 14$\times$14$\times$512 & 3$\times$3$\times$256,1\\
\hline
conv4b & 14$\times$14$\times$512 & 14$\times$14$\times$512 & 3$\times$3$\times$512,1\\
\hline
pool4 & 14$\times$14$\times$512 & 7$\times$7$\times$512 & 2$\times$2$\times$1,2\\
\hline
fc1 & 7$\times$7$\times$512 & 1$\times$1$\times$256 & -\\
\hline
fc2 & 1$\times$1$\times$256 & 1$\times$1$\times$136 & -\\
\hline
\end{tabular}
\end{table}

\subsection{Feed-forward neural network} \label{sec:feed-forward}
The structure of the feed-forward part of each stage is shown in Table \ref{tab:feed-forward}. With the exception of max pooling layers and the output layer, every layer takes advantage of batch normalization and uses Rectified Linear Units (ReLU) for activations. A dropout \cite{dropout} layer is added before the first fully connected layer. The last layer outputs the update $\Delta S_t$ to the current estimate of the landmark positions. 

The overall shape of the feed-forwad network was inspired by the network used in \cite{VGG} for the ImageNet ILSVRC 2014 competition. 

\begin{figure}
\centering
\includegraphics[width=0.24\linewidth]{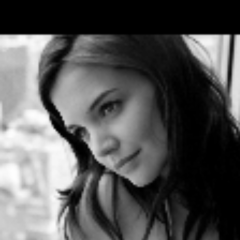}
\includegraphics[width=0.24\linewidth]{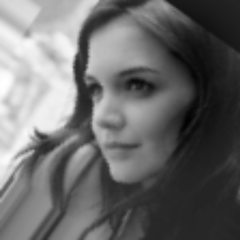}
\includegraphics[width=0.24\linewidth]{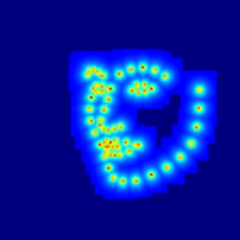}
\includegraphics[width=0.24\linewidth]{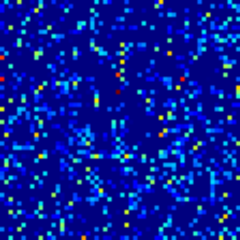}

\includegraphics[width=0.24\linewidth]{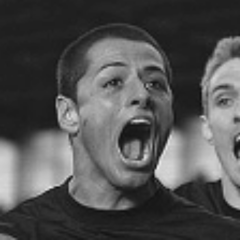}
\includegraphics[width=0.24\linewidth]{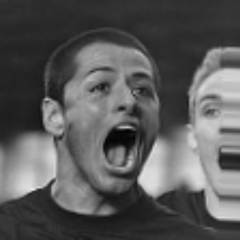}
\includegraphics[width=0.24\linewidth]{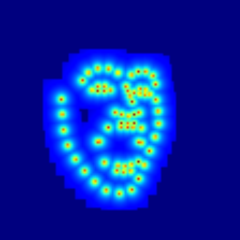}
\includegraphics[width=0.24\linewidth]{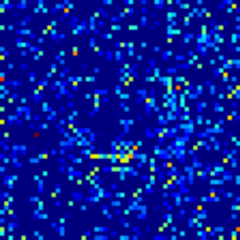}

\includegraphics[width=0.24\linewidth]{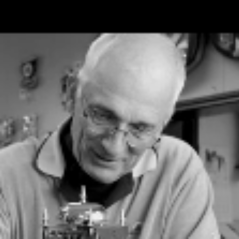}
\includegraphics[width=0.24\linewidth]{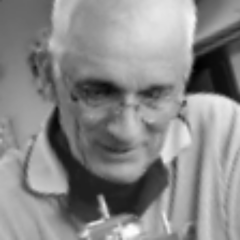}
\includegraphics[width=0.24\linewidth]{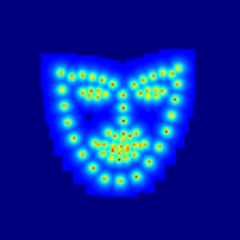}
\includegraphics[width=0.24\linewidth]{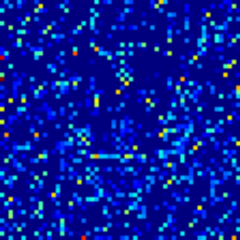}

\caption{Selected images from the IBUG dataset and intermediate results after 1 stage of DAN. The columns show: the input image $I$, the input image normalized to canonical shape using transform $T_2$, the landmark heatmap showing $T_2(S_1)$, the corresponding feature image.}
 \label{fig:afterStage1}
\end{figure}

\subsection{Normalization to canonical shape} \label{sec:normalization}
In DAN the input image $I$ is transformed for each stage so that the current estimates of the landmarks are aligned with the canonical face shape $S_0$. This normalization step allows the further stages of DAN to be invariant to a given family of transforms. This in turn simplifies the alignment task and improves accuracy.

The Transform Estimation layer of our network is responsible for estimating the parameters of transform $T_{t+1}$ at the output of stage $t$. As input the layer takes the output of the current stage $S_{t}$. Once $T_{t+1}$ is estimated the Image Transform and the Landmark Transform layers transform the image $I$ and landmarks $S_{t}$ to the canonical pose. The image is transformed using bilinear interpolation. Note that for the first stage of DAN the normalization step is not necessary since the input shape is always the average face shape $S_0$, which is also the canonical face shape.

Since the input image is transformed, the output of every stage has to be transformed back to match the original image, the output of any DAN stage is thus:
\begin{equation}
S_t = T^{-1}_t(T_t(S_{t-1}) + \Delta S_t),
\end{equation}
where $\Delta S_t$ is the output of the last layer of stage $t$ and $T^{-1}_t$ is the inverse of transform $T_t$

A similar normalization step has been previously proposed in \cite{MIX} with the use of affine transforms. In our implementation we chose to use similarity transforms as they do not cause non-uniform scaling and skewing of the output image. Figure \ref{fig:afterStage1} shows examples of images before and after the transformation. 

\subsection{Landmark heatmap} \label{sec:heatmap}
The landmark heatmap is an image where the intensity is highest in the locations of landmarks and it decreases with the distance to the closest landmark. Thanks to the use of landmark heatmaps the Convolutional Neural Network can infer the landmark locations estimated by the previous stage. In consequence DAN can perform face alignment based on entire facial images. 

At the input to a DAN stage the landmark heatmap is created based on the landmark estimates produced by the previous stage and transformed to the canonical pose: $T_t(S_{t-1})$. The heatmap is generated using the following equation:
\begin{equation}
H(x,y)=\frac{1}{1 + \min_{s_i \in T_t(S_{t-1})} ||(x, y) - s_i||},
\end{equation}
where $H$ is the heatmap image and $s_i$ is the i-th landmark of $T_t(S_{t-1})$. In our implementation the heatmap values are only calculated in a circle of radius 16 around each landmark to improve performance. Note that similarly to normalization, this step is not necessary at the input of the first stage, since the input shape is always assumed to be $S_0$, which would result in an identical heatmap for any input.

An example  of a face image and a corresponding landmark heatmap is shown in Figure \ref{fig:afterStage1}.

\subsection{Feature image layer} \label{sec:feature}
The feature image layer $F_t$ is an image created from a dense layer connected to the fc1 layer (see Table \ref{tab:feed-forward}) of the previous stage $t-1$. Such a connection allows any information learned by the preceding stage to be transferred to the consecutive stage. This naturally complements the heatmap which transfers the knowledge about landmark locations learned by the previous stage.

The feature image layer is a dense layer which has 3136 units with ReLU activations. The output of this dense layer is reshaped to a 56$\times$56 2D layer and upscaled to 112$\times$112, which is the input shape of DAN stages. We use the smaller 56$\times$56 image rather than 112$\times$112 since it showed similar results in our experiments, with considerably less parameters. Figure \ref{fig:afterStage1} shows an example of a feature image.

\subsection{Training procedure} \label{sec:training}
The stages of DAN are trained sequentially. The first stage is trained by itself until the validation error stops improving. Subsequently the connection layers and the second stage are added and trained. This procedure is repeated until further stages stop reducing the validation error.

While many face alignment methods \cite{ESR, SDM, LBF} learn a model that minimizes the Sum of Squared Errors of the landmark locations, DAN minimizes the landmark location error normalized by the distance between the pupils:
\begin{equation} \label{interpupil}
 \min_{\Delta S_t} \frac{||T^{-1}_t(T_t(S_{t-1}) + \Delta S_t) - S^*||}{d_{ipd}}, 
\end{equation}
where $S^*$ is a vector of ground truth landmark locations, $T_t$ is the transform that normalizes the input image and shape for stage $t$ and $d_{ipd}$ is the distance between the pupils of $S^*$. The use of this error is motivated by the fact that it is a far more common \cite{SDM,LBF,CFSS} benchmark for face alignment methods than the Sum of Squared Errors.

Thanks to the fact that all of the layers used in DAN are differentiable DAN can also be trained end-to-end. In order to evaluate end-to-end training in DAN we have experimented with several approaches. Pre-training the first stage for several epochs followed by training of the entire network yielded similar accuracy to the proposed approach but the training was significantly longer. Training the entire network from scratch yielded results significantly inferior to the proposed approach. 

While we did not manage to obtain improved results with end-to-end training we believe that it is possible with a better training strategy. We leave the creation of such a strategy for future work.

\section{Experiments} \label{experiments}
In this section we perform an extensive evaluation of the proposed method on several public datasets as well as the test set of the Menpo challenge \cite{Menpo} to which we have submitted our method. The following paragraphs detail the datasets, error measures and implementation. Section \ref{sec:comparison} compares our method with the state of the art, while section \ref{sec:menpo} shows our results in the Menpo challenge. Section \ref{sec:further} discusses the influence of the number of stages on the performance of DAN.

\paragraph*{Datasets}
In order to evaluate our method we perform experiments on the data released for the 300W competition \cite{300-W} and the recently introduced Menpo challenge \cite{Menpo} dataset. 

The 300W competition data is a compilation of images from five datasets: LFPW \cite{LFPW}, HELEN \cite{HELEN}, AFW \cite{AFW}, IBUG \cite{300-W} and 300W private test set \cite{300-W}. The last dataset was originally used for evaluating competition entries and at that time was private to the organizers of the competition, hence the name. Each image in the dataset is annotated with 68 landmarks \cite{Reannotation} and accompanied by a bounding box generated by a face detector. We follow the most established approach \cite{LBF,CFSS,KRFWS,RAR} and divide the 300-W competition data into training and testing parts. The training part consists of the AFW dataset as well as training subsets of LFPW and HELEN, which results in a total of 3148 images. The test data consists of the remaining datasets: IBUG, 300W private test set, test sets of LFPW, HELEN. In order to facilitate comparison with previous methods we split this test data into four subsets: 

\begin{itemize}
  \item the \textit{common subset} which consists of the test subsets of LFPW and HELEN (554 images),
  \item the \textit{challenging subset} which consists of the IBUG dataset (135 images),
  \item the \textit{300W public test set} which consists of the test subsets of LFPW and HELEN as well as the IBUG dataset (689 images),
  \item the \textit{300W private test set} (600 images).
\end{itemize}

The annotation for the images in the 300W public test set were originally published for the 300W competition as part of its training set. We use them for testing as it became a common practice to do so in the recent years \cite{LBF, CFSS, RAR, cGPRT, KRFWS}.

The Menpo challenge dataset consists of semi-frontal and profile face image datasets. In our experiments we only use the semi-frontal dataset. The dataset consists of training and testing subsets containing 6679 and 5335 images respectively. The training subset consists of images from the FDDB \cite{FDDB} and AFLW \cite{AFLW} datasets. The image were annotated with the same set of 68 landmarks as the 300W competition data but no face detector bounding boxes. The annotations of the test subset have not been released. 

\paragraph*{Error measures}
Several measures of face alignment error for an individual face image have been recently introduced: 
\begin{itemize}
	\item the mean distance between the localized landmarks and the ground truth landmarks divided by the inter-ocular distance (the distance between the outer eye corners) \cite{CFSS, LBF, RAR}, 
    \item the mean distance between the localized landmarks and the ground truth landmarks divided by the inter-pupil distance (the distance between the eye centers) \cite{MDM, 300-W},
    \item the mean distance between the localized landmarks and the ground truth landmarks divided by the diagonal of the bounding box \cite{Menpo}.
\end{itemize}

In our work, we report our results using all of the above measures. 
For evaluating our method on the test datasets we use three metrics: the mean error, the area under the cumulative error distribution curve (AUC$_\alpha$) and the failure rate.

Similarly to \cite{MIX, MDM}, we calculate AUC$_\alpha$ as the area under the cumulative distribution curve calculated up to a threshold $\alpha$, then divided by that threshold. As a result the range of the AUC$_\alpha$ values is always 0 to 1. Following \cite{MDM}, we consider each image with an inter-ocular normalized error of 0.08 or greater as failure and use the same threshold for AUC$_{0.08}$. In all the experiments we test on the full set of 68 landmarks.

\paragraph*{Implementation}
We train two models, DAN which is trained on the training subset of the 300W competition data and DAN-Menpo which is trained on both the above mentioned dataset and the Menpo challenge training set. Data augmentation is performed by mirroring around the Y axis as well as random translation, rotation and scaling, all sampled from normal distributions. During data augmentation a total of 10 images are created from each input image in the training set. 

Both models (DAN and DAN-Menpo) consist of two stages. Training is performed using Theano 0.9.0 \cite{Theano} and Lasagne 0.2 \cite{Lasagne}. For optimization we use Adam stochastic optimization \cite{Adam} with an initial step size of 0.001 and mini batch size of 64. For validation we use a random subset of 100 images from the training set. 

The Python implementation runs at 73 fps for images processed in parallel and at 45 fps for images processed sequentially on a GeForce GTX 1070 GPU. We believe that the processing speed can be further improved by optimizing the implementation of some of our custom layers, most notably the Image Transform layer.

To enable reproducible research, we release the source code of our implementation as well as the models used in the experiments\footnote{\url{https://github.com/MarekKowalski/DeepAlignmentNetwork}}. The published implementation also contains an example of face tracking with the proposed method.

\subsection{Comparison with state-of-the-art} \label{sec:comparison}
We compare the DAN model with state-of-the-art methods on all of the test sets of the 300W competition data. We also show results for the DAN-Menpo model but do not perform comparison since at the moment there are no published methods that use this dataset for training. For each test set we initialize our method using the face detector bounding boxes provided with the datasets.

Tables \ref{tab:public1} and \ref{tab:public2} show the mean error, AUC$_{0.08}$ and the failure rate of the proposed method and other methods on the \textit{300W public test set}. Table \ref{tab:private} shows the mean error, the AUC$_{0.08}$ and failure rate on the \textit{300W private test set}. 

All of the experiments performed on the two most difficult test subsets (the \textit{challenging subset} and the \textit{300W private test set}) show state-of-the-art results, including:
\begin{itemize}
	\item a failure rate reduction of 60\% on the \textit{300W private test set}, 
    \item a failure rate reduction of 72\% on the  \textit{300W public test set},
    \item a 9\% improvement of the mean error on the \textit{challenging subset}.
\end{itemize}
This shows that the proposed DAN is particularly suited for handling difficult face images with a high degree of occlusion and variation in pose and illumination.

\begin{table}[tb]
\caption{Mean error of face alignment methods on the 300W public test set and its subsets. All values are shown as percentage of the normalization metric.} \label{tab:public1}
\begin{tabularx}{\linewidth}{ >{\centering\arraybackslash}X c c c c }
\Xhline{4\arrayrulewidth}
Method & \makecell{Common \\ subset} & \makecell{Challenging \\ subset} & Full set \\
\hline
\multicolumn{4}{c}{inter-pupil normalization} \\
\hline
ESR \cite{ESR} & 5.28 & 17.00 & 7.58\\
SDM \cite{SDM} & 5.60 & 15.40 & 7.52\\
LBF \cite{LBF} & 4.95 & 11.98 & 6.32\\
cGPRT \cite{cGPRT} & - & - & 5.71\\
CFSS \cite{CFSS} & 4.73 & 9.98 & 5.76\\
Kowalski et al. \cite{KRFWS} & 4.62 & 9.48 & 5.57\\
RAR \cite{RAR} & \textbf{4.12} & 8.35 & 4.94\\

\hline	
\textbf{DAN} & 4.42& \textbf{7.57} & 5.03 & \\
\textbf{DAN-Menpo} & 4.29 & \textbf{7.05} & \textbf{4.83} & \\
\hline
\multicolumn{4}{c}{inter-ocular normalization} \\
\hline
MDM \cite{MDM} & - & - & 4.05\\
Kowalski et al. \cite{KRFWS} & 3.34 & 6.56 & 3.97\\

\hline	
\textbf{DAN} & \textbf{3.19} & \textbf{5.24} & \textbf{3.59} & \\
\textbf{DAN-Menpo} & \textbf{3.09} & \textbf{4.88} & \textbf{3.44} & \\

\hline
\multicolumn{4}{c}{bounding box diagonal normalization} \\
\hline
\textbf{DAN} & \textbf{1.35} & \textbf{2.00} & \textbf{1.48} & \\
\textbf{DAN-Menpo} & \textbf{1.31} & \textbf{1.87} & \textbf{1.42} & \\

\Xhline{4\arrayrulewidth}
\end{tabularx}
\end{table}

\begin{table}[tb]
\caption{AUC and failure rate of face alignment methods on the 300W public test set. } \label{tab:public2}
\begin{tabularx}{\linewidth}{ >{\centering\arraybackslash}X c c c c }
\Xhline{4\arrayrulewidth}
Method & AUC$_{0.08}$ & Failure (\%) \\
\hline
\multicolumn{3}{c}{inter-ocular normalization} \\
\hline
ESR \cite{ESR} & 43.12 & 10.45\\
SDM \cite{SDM} & 42.94 & 10.89\\
CFSS \cite{CFSS} & 49.87 & 5.08\\
MDM \cite{MDM} & 52.12 & 4.21\\
\hline	
\textbf{DAN} & \textbf{55.33} & \textbf{1.16} & \\
\textbf{DAN-Menpo} & \textbf{57.07} & \textbf{0.58} & \\

\Xhline{4\arrayrulewidth}
\end{tabularx}
\end{table}

\begin{table}[tb]
\caption{Results of face alignment methods on the 300W private test set. Mean error is shown as percentage of the inter-ocular distance.} \label{tab:private}
\begin{tabularx}{\linewidth}{ >{\centering\arraybackslash}X c c c c }
\Xhline{4\arrayrulewidth}
Method & Mean error & AUC$_{0.08}$  & Failure (\%) \\
\hline
\multicolumn{4}{c}{inter-ocular normalization} \\
\hline
ESR \cite{ESR} & - & 32.35 & 17.00\\
CFSS \cite{CFSS} & - & 39.81 & 12.30\\
MDM \cite{MDM} & 5.05 & 45.32 & 6.80\\
\hline	
\textbf{DAN} & \textbf{4.30} & \textbf{47.00} & \textbf{2.67} & \\
\textbf{DAN-Menpo} & \textbf{3.97} & \textbf{50.84} & \textbf{1.83} & \\

\Xhline{4\arrayrulewidth}
\end{tabularx}
\end{table}

\begin{figure}[!htb]
\centering
\includegraphics[width=0.325\linewidth]{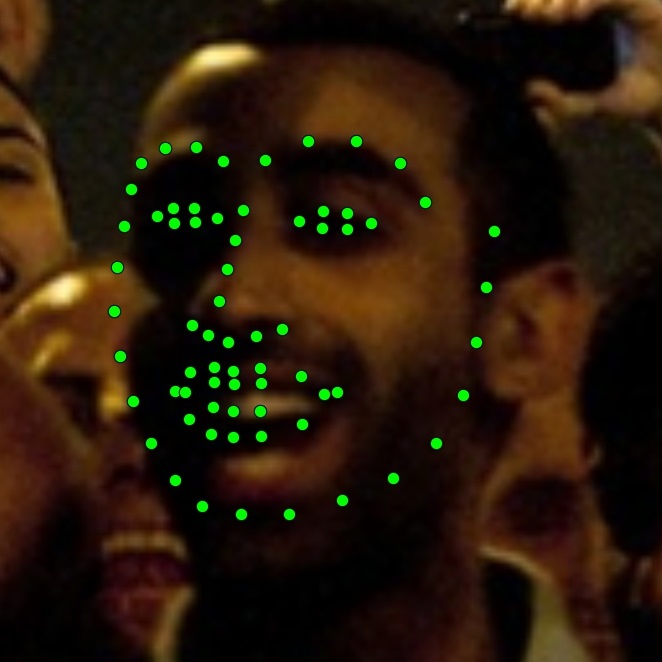}
\includegraphics[width=0.325\linewidth]{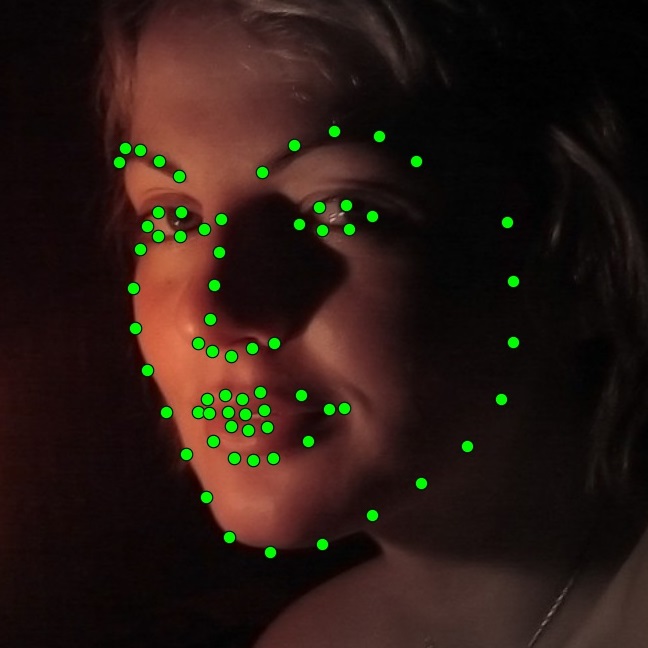}
\includegraphics[width=0.325\linewidth]{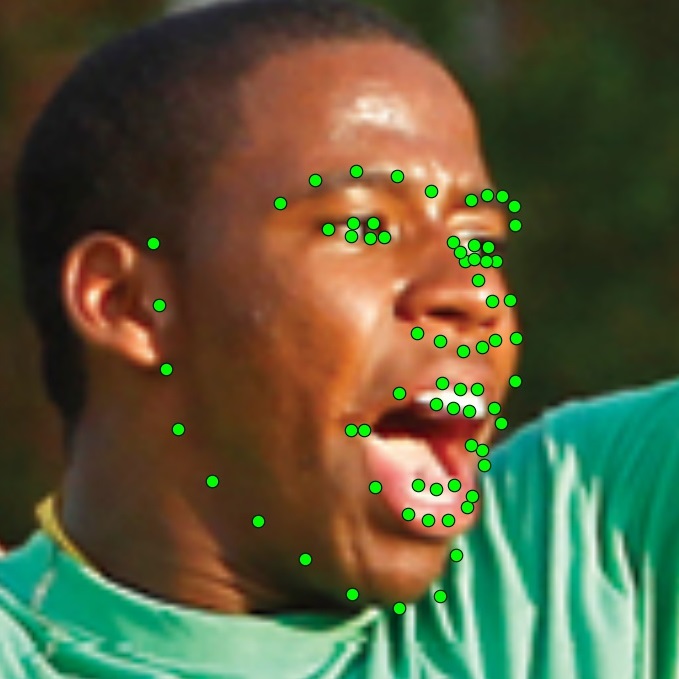}

\includegraphics[width=0.325\linewidth]{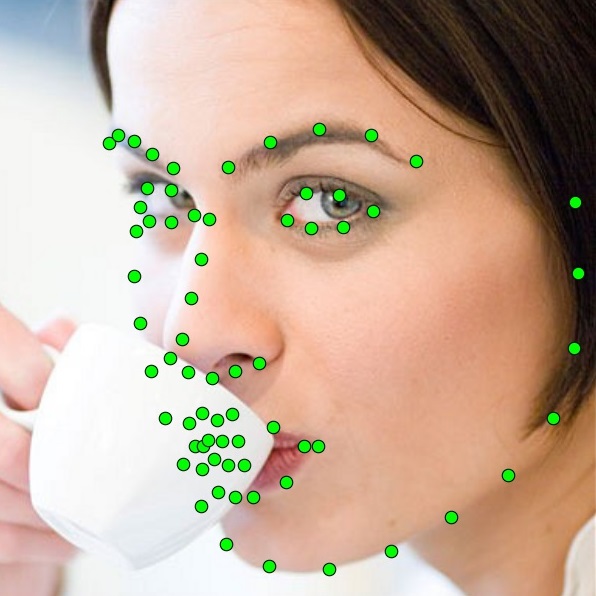}
\includegraphics[width=0.325\linewidth]{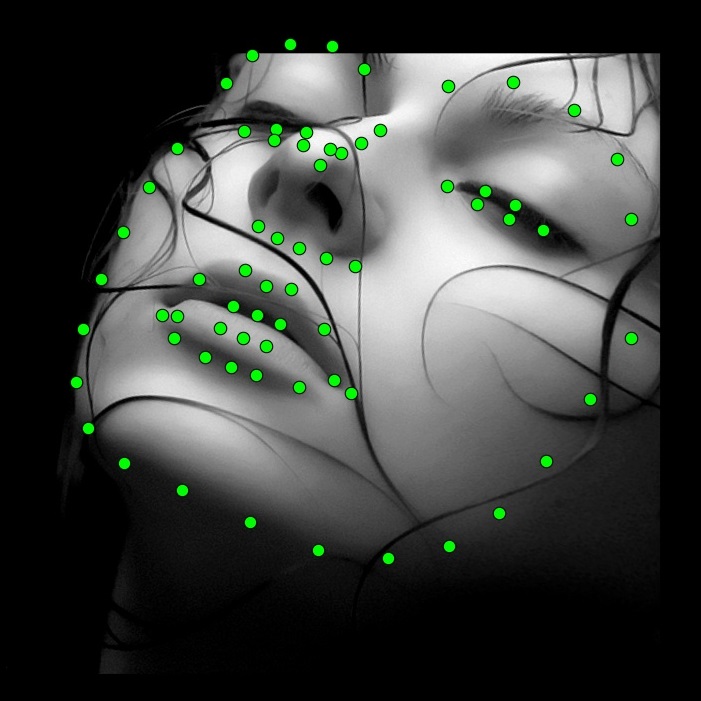}
\includegraphics[width=0.325\linewidth]{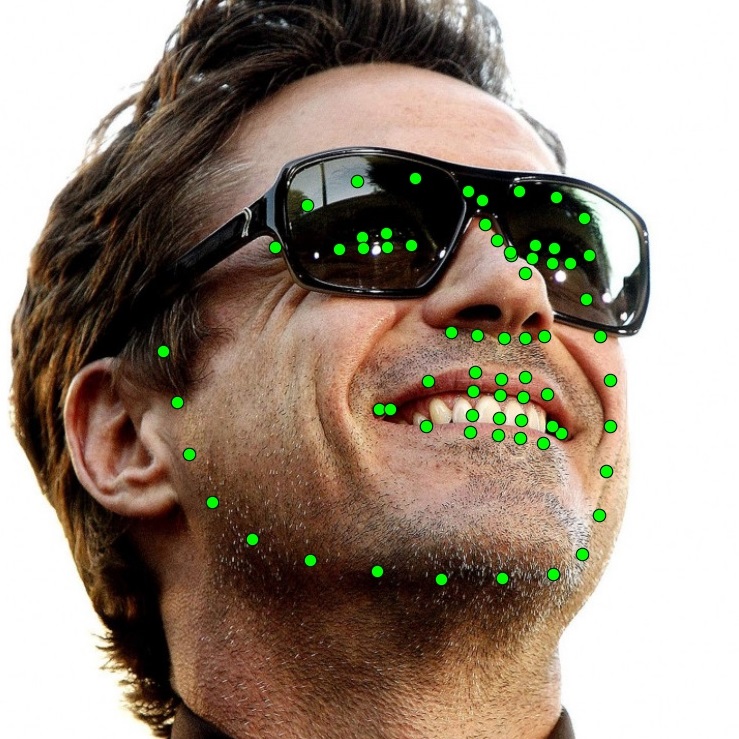}

\includegraphics[width=0.325\linewidth]{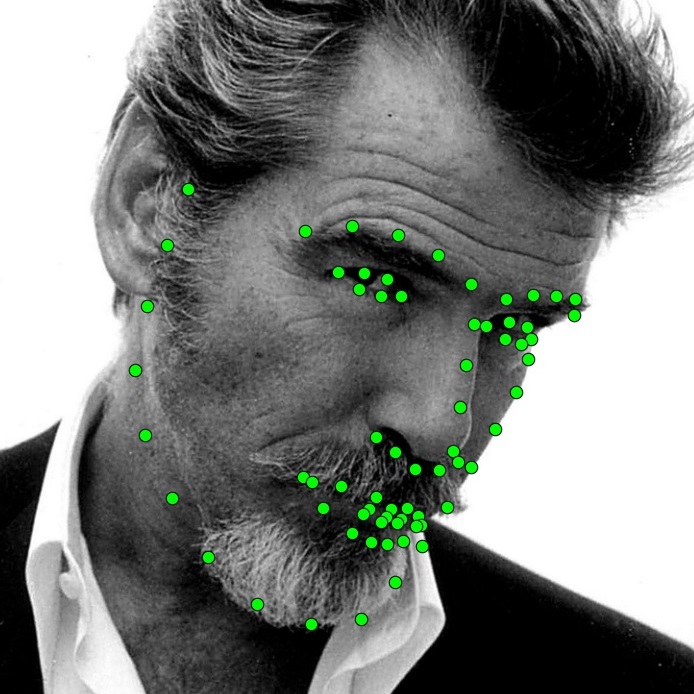}
\includegraphics[width=0.325\linewidth]{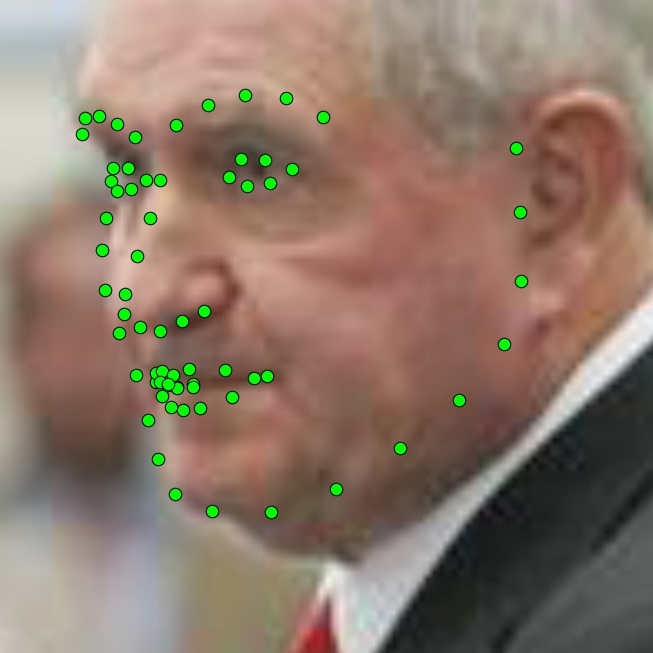}
\includegraphics[width=0.325\linewidth]{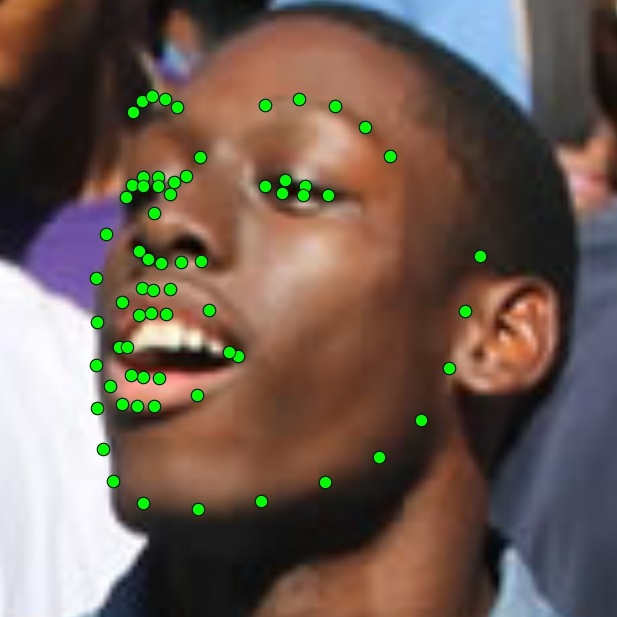}
\caption{ The 9 \textit{worst} results on the challenging subset (IBUG dataset) in terms of inter-ocular error produced by the DAN model. Only the first 7 images have an error of more than 0.08 inter-ocular distance and can be considered failures. }
 \label{fig:ibug_worst}
\end{figure} 

\begin{figure*}[!htb]
\centering
\includegraphics[width=0.16\textwidth]{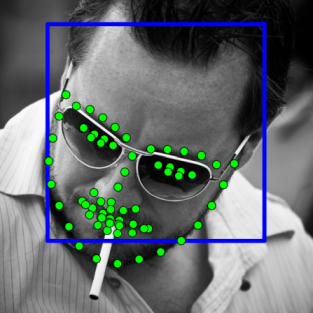}
\includegraphics[width=0.16\textwidth]{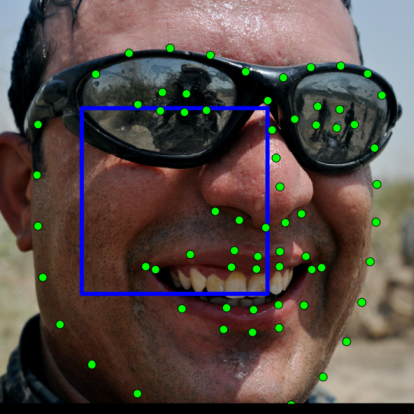}
\includegraphics[width=0.16\textwidth]{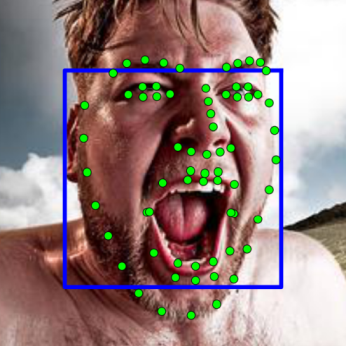}
\includegraphics[width=0.16\textwidth]{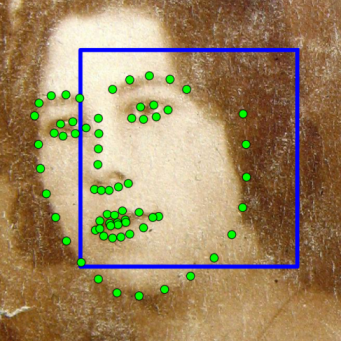}
\includegraphics[width=0.16\textwidth]{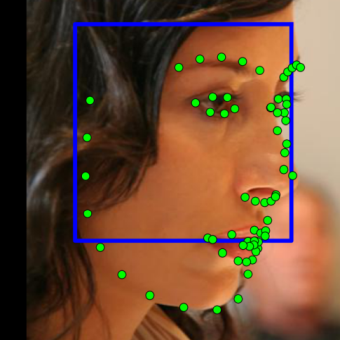}
\includegraphics[width=0.16\textwidth]{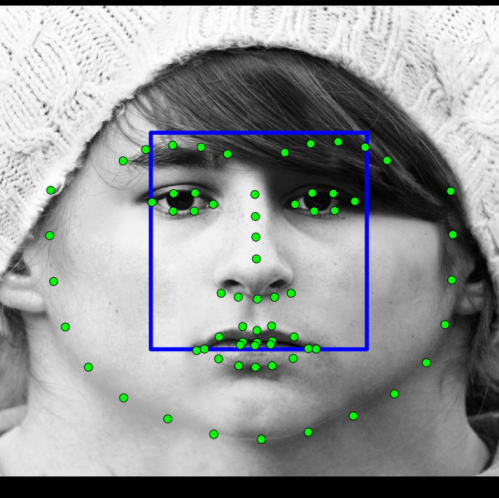}

\includegraphics[width=0.16\textwidth]{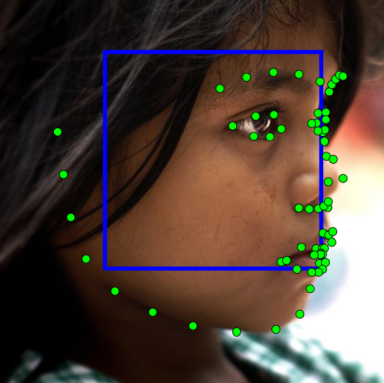}
\includegraphics[width=0.16\textwidth]{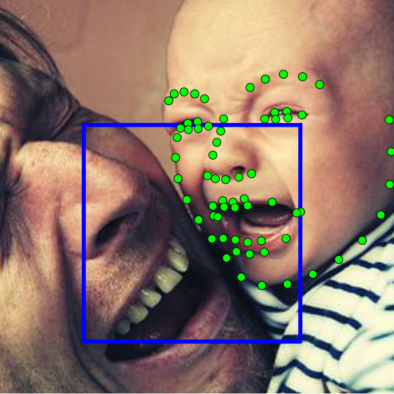}
\includegraphics[width=0.16\textwidth]{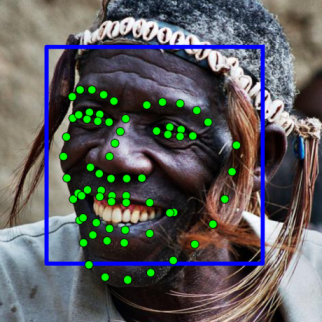}
\includegraphics[width=0.16\textwidth]{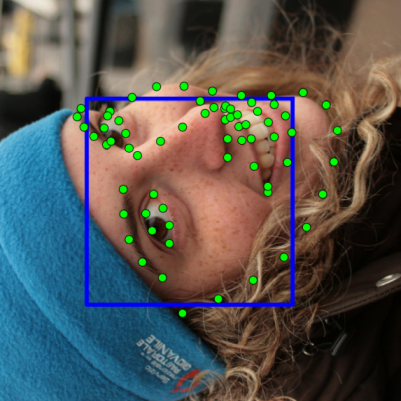}
\includegraphics[width=0.16\textwidth]{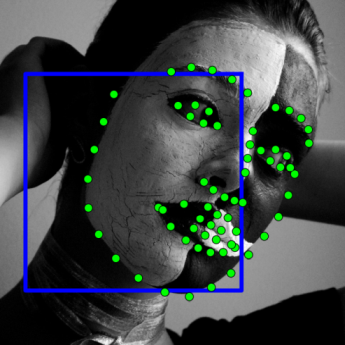}
\includegraphics[width=0.16\textwidth]{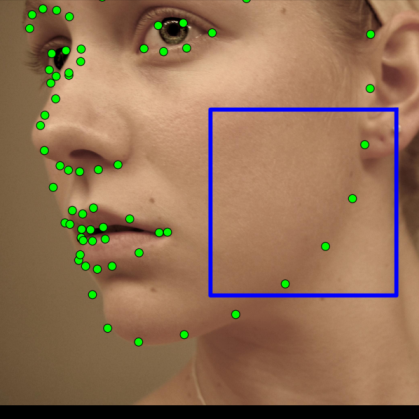}

\caption{ Results of our submission to the Menpo challenge on some of the difficult images of the Menpo test set. The blue squares denote the initialization bounding boxes. The images were cropped to better visualize the results, in the original images the bounding boxes are always located in the center. } \label{fig:menpo_examples}
\end{figure*}

\subsection{Results on the Menpo challenge test set} \label{sec:menpo}
In order to evaluate the proposed method on the Menpo challenge test dataset we have submitted our results to the challenge and received the error scores from the challenge organizers. The Menpo test data differs from the other datasets we used in that it does not include any bounding boxes which could be used to initialize face alignment. For that reason we have decided to use a two step face alignment procedure, where the first step serves as an initialization for the second step.

The first step performs face alignment using a square initialization bounding box placed in the middle of the image with a size set to a percentage of image height. The second step takes the result of the first step, transforms the landmarks and the image to the canonical face shape and creates a bounding box around the transformed landmarks. The transformed image and bounding box are used as input to face alignment. An inverse transform is later applied to get landmark coordinates for the original image. 

In order to determine the optimal size of the bounding boxes in the first step we ran DAN on a small subset of the Menpo test set for several bounding box sizes. The optimal size was determined using a method that would estimate the face alignment error of a given set of landmarks and an image. Said method extracts HOG \cite{HOG} features at each of the landmarks and uses a linear model to estimate the error. The method was trained on the 300W training set using ridge regression. The chosen bounding box size was 46\% of the image height.

\begin{figure}[!htb]
\centering
\includegraphics[width=\linewidth]{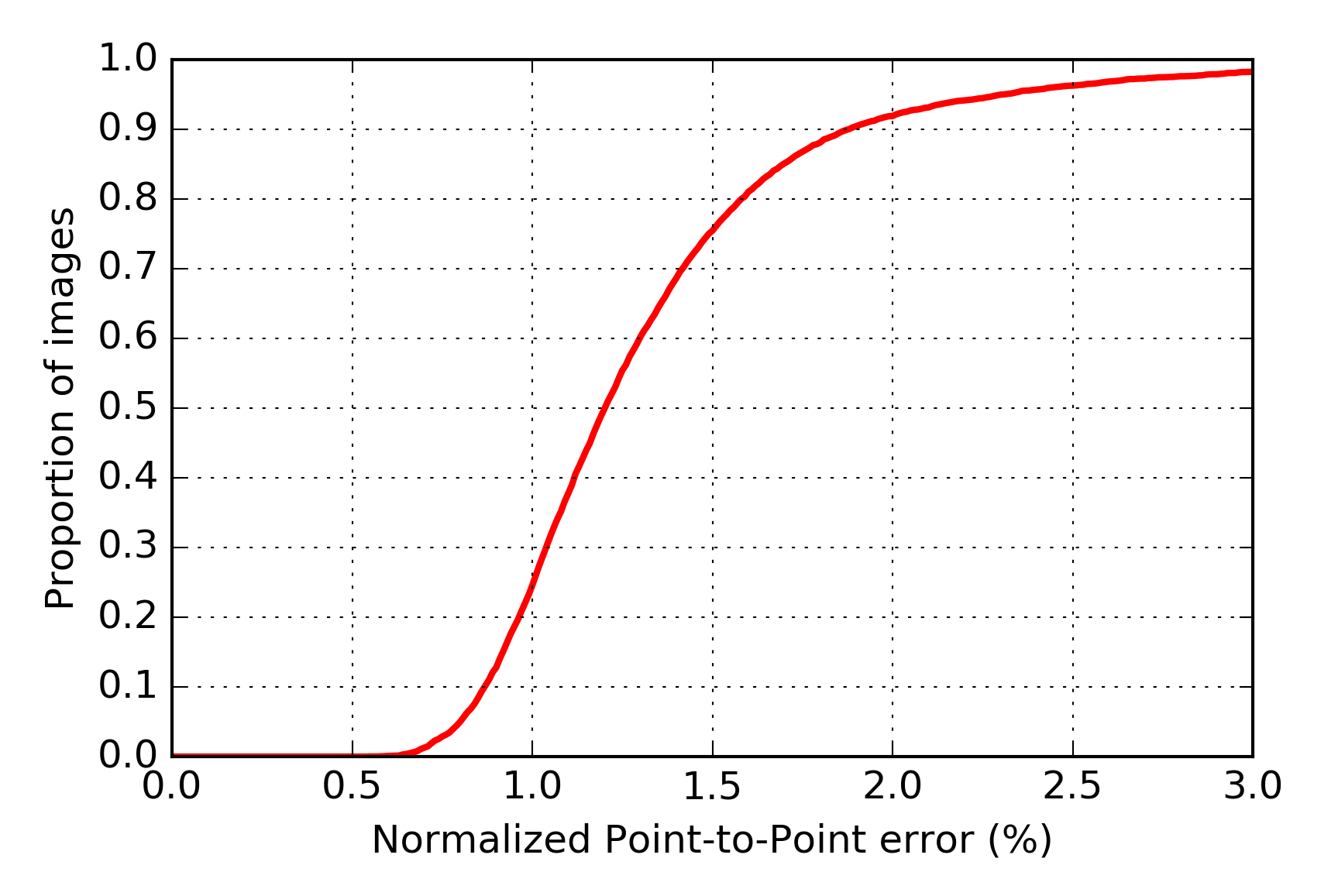}
\caption{The Cumulative Error Distribution curve for the DAN-Menpo model on the Menpo test set. The Point-to-Point error is shown as percentage of the bounding box diagonal.}
 \label{fig:Menpo}
\end{figure}

Figure \ref{fig:Menpo} and Table \ref{tab:Menpo} show the CED curve, mean error, $AUC_{0.03}$ and failure rate for the DAN-Menpo model on the Menpo test set. In all cases the errors are calculated using the diagonal of the bounding box normalization, used by the challenge organizers. For the AUC and the failure rate we have chosen a threshold of 0.03 of the bounding box diagonal as it is approximately equivalent to 0.08 of the interocular distance used in the previous chapter.
 
Figure \ref{fig:menpo_examples} shows examples of images from the Menpo test set and corresponding results produced by our method. Note that even though DAN was trained primarily on semi-frontal images it can handle fully profile images as well.

\begin{table}[tb]
\caption{Results of the proposed method on the semi-frontal subset of the Menpo test set. Mean error is shown as percentage of the bounding box diagonal.} \label{tab:Menpo}
\begin{tabularx}{\linewidth}{ >{\centering\arraybackslash}X c c c c }
\Xhline{4\arrayrulewidth}
Method & Mean error & AUC$_{0.03}$  & Failure (\%) \\
\hline
\multicolumn{4}{c}{bounding box diagonal normalization} \\
\hline
\textbf{DAN-Menpo} & \textbf{1.38} & \textbf{56.20} & \textbf{1.74} & \\
\Xhline{4\arrayrulewidth}
\end{tabularx}
\end{table}

\subsection{Further evaluation} \label{sec:further}
In this subsection we evaluate several DAN models with a varying number of stages on the 300W private test set. All of the models were trained identically to the DAN model from section \ref{sec:comparison}.
Table \ref{tab:stage_number} shows the results of our evaluation. The addition of the second stage increases the AUC$_{0.08}$ by 20\% while the mean error and failure rate are reduced by 14\% and 56\% respectively. The addition of a third stage does not bring significant benefit in any of the metrics.

\begin{table}[tb]
\caption{Results of the proposed method with a varying number of stages on the 300W private test set. Mean error is shown as percentage of the inter-ocular distance.} \label{tab:stage_number}
\begin{tabularx}{\linewidth}{ >{\centering\arraybackslash}X c c c c }
\Xhline{4\arrayrulewidth}
\# of stages & Mean error & AUC$_{0.08}$  & Failure (\%) \\
\hline
\multicolumn{4}{c}{inter-ocular normalization} \\
\hline
1 & 5.02 & 39.04 & 6.17 & \\
2 & 4.30 & 47.00 & 2.67 & \\
3 & 4.32 & 47.08 & 2.67 & \\
\Xhline{4\arrayrulewidth}
\end{tabularx}
\end{table}

\section{Conclusions}
In this paper, we introduced the Deep Alignment Network - a robust face alignment method based on convolutional neural networks. Contrary to the recently proposed face alignment methods, DAN performs face alignment based on entire face images, which makes it highly robust to large variations in both initialization and head pose. Using entire face images instead of local patches extracted around landmarks is possible thanks to the use of novel landmark heatmaps which transmit the information about landmark locations between DAN stages. Extensive evaluation performed on two challenging, publicly available datasets shows that the proposed method improves the state-of-the-art failure rate by a significant margin of over 70\%.

Future research includes investigation of new strategies for training DAN in an end-to-end manner. We also plan to introduce learning into the estimation of the transform $T_t$ that normalizes the shapes and images between stages.

\section{Acknowledgements}
The work presented in this article was supported by The National Centre for Research and Development grant number DOB-BIO7/18/02/2015. The results obtained in this work were a basis for developing the software for the grant sponsor, which is different from the software published with this paper.

We thank NVIDIA for donating a Titan X Pascal GPU which was used to train the proposed neural network.

{\small
\bibliographystyle{ieee}
\bibliography{egbib}
}

\end{document}